\documentclass[letterpaper, 10 pt, conference]{ieeeconf}
\IEEEoverridecommandlockouts
\usepackage{amsmath,amsfonts}
\usepackage{algorithmic}
\usepackage{algorithm}
\usepackage{array}
\usepackage{textcomp}
\usepackage{stfloats}
\usepackage{url}
\usepackage{verbatim}
\usepackage{multirow}
\usepackage{cite}
\usepackage{hyperref}
\hypersetup{
    colorlinks=true,
    urlcolor=blue
}
\usepackage{pgfplots}
\pgfplotsset{compat=1.17}

\usepackage{tikzpagenodes}
\usepackage{atbegshi} 

\AtBeginShipout{%
  \ifnum\value{page}=1
    \AtBeginShipoutUpperLeft{%
      \begin{tikzpicture}[remember picture,overlay]
        \node[anchor=south,yshift=5mm,text width=\paperwidth,align=center] 
          at (current page.south)
          {\footnotesize
          © 2024 IEEE. This is the accepted version of the paper. Rößle, D., Gerner, J., Bogenberger, K., Cremers, D., Schmidtner, S., \& Schön, T.\\
          \textit{Unlocking Past Information: Temporal Embeddings in Cooperative Bird’s Eye View Prediction}. In 2024 IEEE Intelligent Vehicles Symposium (IV), pp. 2220–2225, 2024. IEEE. \href{https://doi.org/10.1109/IV55156.2024.10588608}{https://doi.org/10.1109/IV55156.2024.10588608}
          };
      \end{tikzpicture}
    }%
  \fi
}

\title{\LARGE \bf Unlocking Past Information: Temporal Embeddings in Cooperative Bird's Eye View Prediction}

\author{Dominik Rößle$^{1}$, Jeremias Gerner$^{2}$, Klaus Bogenberger$^{3}$, Daniel Cremers$^{4}$, \\ Stefanie Schmidtner$^{5}$, Torsten Schön$^{6}$
\thanks{$^{1}$Dominik Rößle is with the Department of Computer Science and AImotion Bavaria, Technische Hochschule Ingolstadt, 85049 Ingolstadt, Germany
        {\tt\small dominik.roessle@thi.de}}
\thanks{$^{2}$Jeremias Gerner is with the Department of Electrical Engineering and AImotion Bavaria, Technische Hochschule Ingolstadt, 85049 Ingolstadt, Germany
        {\tt\small jeremias.gerner@thi.de}}
\thanks{$^{3}$Klaus Bogenberger is with the School of Engineering and Design, Technical University of Munich, 80333 München, Germany
        {\tt\small klaus.bogenberger@tum.de}}
\thanks{$^{4}$Daniel Cremers is with the School of Computation, Information and Technology, Technical University of Munich, 85748 Garching, Germany
        {\tt\small cremers@tum.de}}
\thanks{$^{5}$Stefanie Schmidtner is with the Department of Electrical Engineering and AImotion Bavaria, Technische Hochschule Ingolstadt, 85049 Ingolstadt, Germany
        {\tt\small stefanie.schmidtner@thi.de}}
\thanks{$^{6}$Torsten Schön is with the Department of Computer Science and AImotion Bavaria, Technische Hochschule Ingolstadt, 85049 Ingolstadt, Germany
        {\tt\small torsten.schoen@thi.de}}
}

\begin{document}

\maketitle
\thispagestyle{empty}
\pagestyle{empty}

\begin{abstract}
Accurate and comprehensive Bird's Eye View (BEV) semantic segmentation is essential for ensuring safe and proactive navigation in autonomous driving.
Although cooperative perception has exceeded the detection capabilities of single-agent systems, prevalent camera-based algorithms in cooperative perception neglect valuable information derived from historical observations.
This limitation becomes critical during sensor failures or communication issues as cooperative perception reverts to single-agent perception, leading to degraded performance and incomplete BEV segmentation maps.
This paper introduces TempCoBEV, a temporal module designed to incorporate historical cues into current observations, thereby improving the quality and reliability of BEV map segmentations.
We propose an importance-guided attention architecture to effectively integrate temporal information that prioritizes relevant properties for BEV map segmentation.
TempCoBEV is an independent temporal module that seamlessly integrates into state-of-the-art camera-based cooperative perception models.
We demonstrate through extensive experiments on the OPV2V dataset that TempCoBEV performs better than non-temporal models in predicting current and future BEV map segmentations, particularly in scenarios involving communication failures.
We show the efficacy of TempCoBEV and its capability to integrate historical cues into the current BEV map, improving predictions under optimal communication conditions by up to 2\% and under communication failures by up to 19\%.
The code is available at \url{https://github.com/cvims/TempCoBEV}.
\end{abstract}

\section{Introduction}
Autonomous driving relies on a precise perception of the driving environment, enabling the vehicle to interpret and respond to complex scenarios, ensuring safety \cite{Zhang2023CoopImpactReview} and proactive navigation.
In addressing precise perception, vehicles often integrate camera and LiDAR systems independently or combined \cite{Yang2023BEVFormerV2, Huang2021bevdet, Liu2023BEVFusion}.
Despite the presence of multiple sensors, accurately perceiving the surroundings of the ego vehicle remains challenging due to unavoidable occlusions \cite{Xu2022OPV2V, gerner2023sumo}, sensor failures \cite{Roessle2022PHP}, or sensor contamination \cite{Wang2022CoopReview}. 
In response to these challenges, collaborative perception has become a compelling solution in recent years \cite{Chen2019FCooper, Wang2020V2VNet, Hu2022Where2comm}.

While LiDAR is a valuable sensor in autonomous driving, it falls behind camera-based solutions regarding cost, resolution, color information, and performance in adverse weather conditions.
Additionally, collaborative perception reveals that camera-based solutions can outperform LiDAR-based approaches \cite{Hu2023CameraOvertakesLiDAR}.
Capitalizing the strengths of camera perspectives, mapping them onto the comprehensive Bird's Eye View (BEV) space provides distinct advantages, preserving both spatial and temporal dimensions of road elements \cite{xu2022cobevt}.
This becomes crucial for various autonomous driving tasks, including scene understanding and planning \cite{Ng2020BEVSeg, Zhou2022CrossViewTransformer}.
However, current methodologies for camera-based BEV map segmentation still rely on single-frame prediction \cite{xu2022cobevt}, neglecting historical context in BEV prediction and failing to consider significant semantic cues embedded in the temporal context.
Moreover, accessing historical information is crucial as it potentially decreases the transmission frequency of Connected and Autonomous Vehicles (CAVs) and acts as a compensatory mechanism for potential communication failures, enhancing overall system reliability and performance.

\begin{figure*}[ht]
  \centering
  \includegraphics[width=\textwidth]{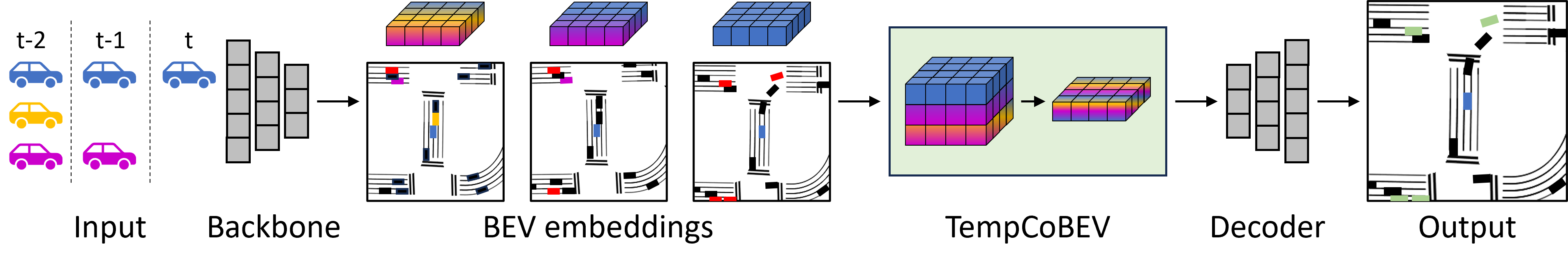}
  \caption{\textbf{Illustration of the TempCoBEV integration.} Different vehicles are shown in different colors. At each timestamp, a varying number of CAVs engage in information sharing. In the illustration of the BEV embeddings, the ego vehicle is shown in blue and is always centered; undetectable vehicles are represented in red, while detectable ones are shown in black. TempCoBEV incorporates current and historical processed embeddings, fusing them into a unified representation before feeding them into the decoder. The resulting output depicts the potential reconstruction of vehicles in green, leveraging historical cues.}
  \label{fig:teaser}
\end{figure*}

In this paper, we introduce \textit{TempCoBEV}, a novel temporal module that leverages fused features from collaborating vehicles to generate BEV segmented maps enriched with temporal cues, as illustrated in Figure \ref{fig:teaser}.
Our module represents an extension for cooperative perception algorithms, augmenting their prediction reliability by incorporating information from past frames.
From the perspective of the ego vehicle, all collaborating vehicles share their intermediate embeddings, which the fusion model then processes into a single fused BEV embedding.
Our temporal module utilizes the current BEV embedding and integrates past cues to enhance the prediction reliability while maintaining contextual continuity in the event of communication failures.

Our main contributions are:
\begin{itemize}
    \item To the best of our knowledge, we are the first to consider developing a temporal module based on the fused embedding level without changing the overall model architecture. TempCoBEV effectively fuses current and historical BEV embeddings, enhancing the prediction reliability by compensating for communication failures.
    \item Our module can be integrated without re-training the initial model architecture and saves up to 24x the training time compared to training the complete model.
    \item We present an importance-guided attention stack to prioritize crucial areas from the historical context, thereby mitigating communication failures. Comprehensive experimental results and ablation studies verify the performance and efficiency of restoring past information.
\end{itemize}

\section{Related Work}
\subsection{Cooperative Perception}
Recent research has primarily concentrated on aggregating features generated by neighboring CAVs.
Initially, due to the absence of a publicly available real-world cooperative driving dataset, Xu et al. \cite{Xu2022OPV2V} introduced the OPV2V dataset, created using CARLA \cite{Dosovitskiy2017Carla}.
They additionally proposed a model with attention-based fusion for feature aggregation among CAVs.
F-Cooper, presented by Chen et al. \cite{Chen2019FCooper}, utilizes a max-pooling operation to aggregate voxel features from CAVs.
Wang et al. \cite{Wang2020V2VNet} introduced V2VNet, employing a spatially aware graph neural network for information aggregation.
Li et al. \cite{Li2021DiscoNet} introduced a student-teacher framework where the early-collaboration teacher model guides the intermediate-collaboration student model, aligning student-created features with the teachers.
All these models use LiDAR data for cooperative perception.
CoBEVT \cite{xu2022cobevt} was the first to use a camera-based cooperative perception, incorporating axial attention to capture sparse spatial semantics.
In this paper, we adapt some of these models for camera-only inputs and integrate our proposed TempCoBEV module to incorporate historical cues, optimizing BEV semantic map predictions.

\subsection{Temporal Integration}
In autonomous driving tasks, incorporating historical information has been widely adopted to address challenges posed by unresolved occlusions or incomplete scenes in single-frame perception.
Li et al. \cite{Li2022BEVFormer} introduced BEVFormer, which employs a spatial cross-attention mechanism based on deformable attention \cite{Zhu2021DeformableAttention} to efficiently integrate the current frame with temporal cues.
Yang et al. \cite{Yang2023SCOPE} presented SCOPE, the first cooperative perception model to integrate temporal information.
SCOPE utilizes temporal data obtained from the ego vehicle and combines it with current data received from surrounding CAVs.
However, this approach incorporates temporal cues solely from the ego perspective, overlooking the perception of CAVs in historical frames, resulting in incomplete traffic perception.
In contrast to prior methods, we focus on camera-based data and consider the temporal context from historical communications with CAVs.

\section{Methodology}
In this paper, we introduce a novel temporal module designed to enhance the perception capabilities of an ego vehicle in collaborative perception scenarios by leveraging historical information.
We aim to enhance the perceptual reliability of detecting dynamic objects and reduce the immediate drop in performance when the connection to previously available CAVs is cut off.
Figures \ref{fig:confidence_fusion} and \ref{fig:temporal_fusion} depict the architecture of the proposed module, comprising three modules: 1) Importance Fusion (\ref{subsec:confidence_estimator}), 2) Temporal Fusion Module (\ref{subsec:temporal_module}), and 3) Feature Aggregation.
Our temporal module is designed as an independent entity, capable of being seamlessly integrated into any camera-based cooperative perception model, showcasing great capabilities in recovering and integrating historical cues into current BEV map segmentations.
In the embedding fusion process, the module assesses the significance of current and historical embeddings by integrating importance scores.
These relative importance scores collectively create an importance map that guides the attention stack module.
This strategic focus enables the model to prioritize the most crucial features in the scene, thereby improving the precision and robustness of the BEV segmentation prediction.
We present results on the OPV2V dataset \cite{Xu2022OPV2V} for the cooperative BEV map segmentation task, employing our module alongside several state-of-the-art models.
Our module effectively utilizes historical information, surpassing the performance of single-frame cooperative models in predicting future frames in communication failure scenarios.

\subsection{Temporal Enhancement in Cooperative Perception for BEV Map Segmentation}
State-of-the-art models for BEV map segmentation commonly incorporate an encoder to process the information of its sensor inputs, generating a BEV embedding ($\mathcal{E}$) that encapsulates crucial details about the position and size of relevant objects.
Within the domain of cooperative perception, BEV embeddings from CAVs are exchanged and fused into a unified BEV embedding.
This fusion process ideally addresses communication delays and positioning discrepancies.
Before decoding the fused embedding, our introduced temporal module is integrated.
Our module takes the fused BEV embeddings from the cooperative perception model as input, enriches them with historical embeddings, and enhances the overall embedding quality.
After the BEV embedding update, the decoder from the original architecture utilizes the enriched representation to generate the corresponding BEV map.
Notably, the fundamental architecture of the cooperative perception model remains unchanged during training.
Since we do not optimize the original fusion models, we pre-infer and store the fused BEV embeddings as a separate dataset.
Subsequently, we drop the backbone, freeze the decoder, and initiate the training of the temporal module using the new embedding dataset as the models' input to speed up training.
Therefore, training focuses solely on the temporal module, leading to time savings of up to 24x compared to training the complete cooperative models with CAV images as input.

\subsection{Importance Fusion}\label{subsec:confidence_estimator}
\begin{figure}[ht]
  \centering
  \includegraphics[width=0.37\textwidth]{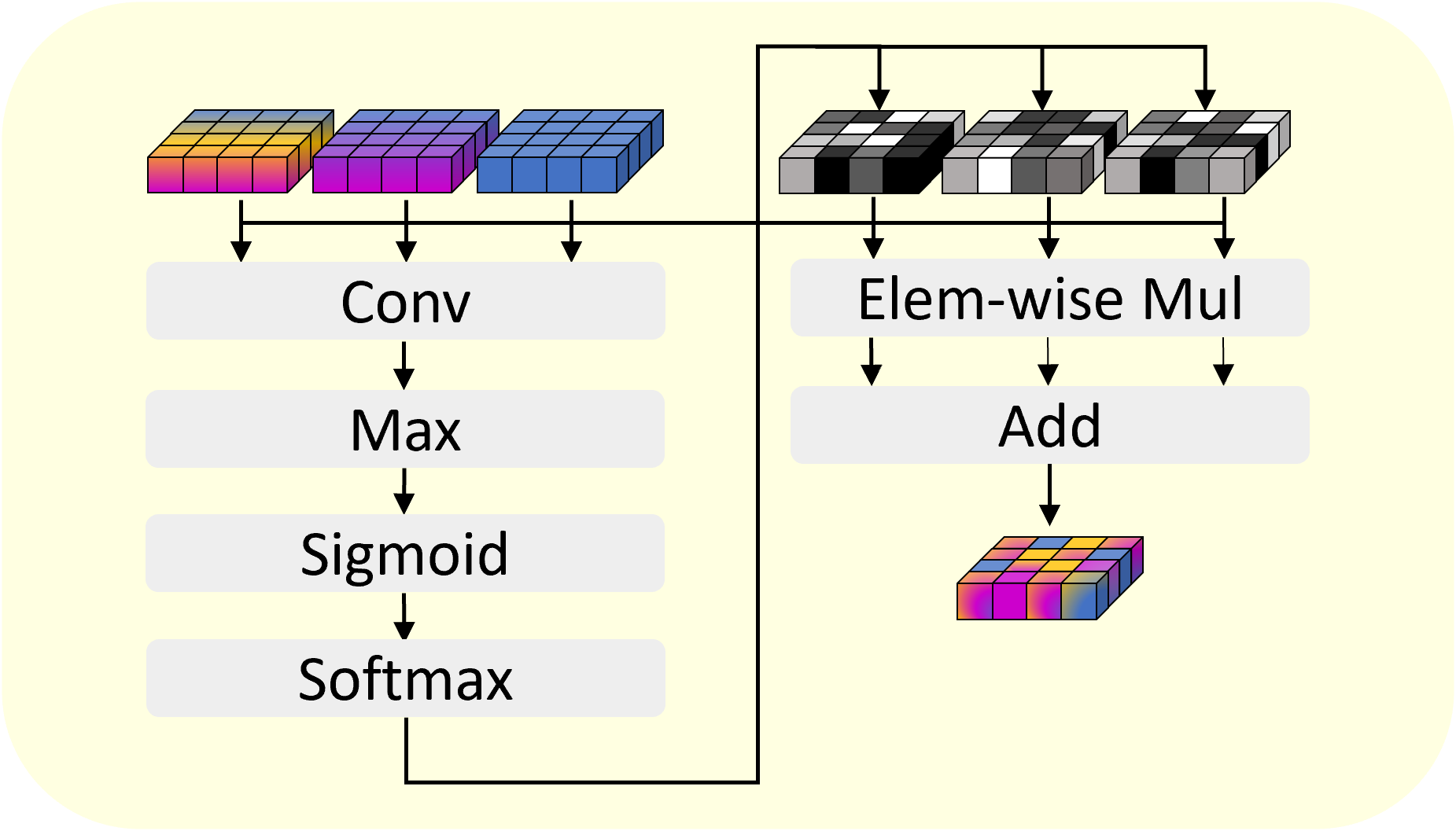}
  \caption{Architecture of the Importance Fusion module to predict importance maps of embeddings and synthesizes information with relative importance.}
  \label{fig:confidence_fusion}
\end{figure}

The primary goal of the importance fusion is to determine the locations of interest and their corresponding importance within a set of $N$ provided embeddings $\mathcal{E}_1, \mathcal{E}_2, \dots, \mathcal{E}_N$, as depicted in Fig. \ref{fig:confidence_fusion}.
The importance estimator follows the importance-aware adaptive fusion module of SCOPE \cite{Yang2023SCOPE} and employs an importance generator $f_{gen}(\cdot)$, which takes the input channels of the BEV embedding and generates $O$ output channels, maintaining the same height and width as the input embedding.
To create the importance map $\mathcal{M}_1$ of an embedding $\mathcal{E}_1$, we use the max value function $\Psi_m$ across the output channels $O$ and apply the sigmoid ($\sigma$).
The importance map of an embedding is calculated as follows:
\begin{equation}
\mathcal{M}_1=\sigma(\Psi_{m}(f_{gen}(\mathcal{E}_1)))
\label{eq:conf_importance}
\end{equation}
The importance maps are then concatenated and put into a softmax function $\phi(\cdot)$ to get the relative importance:
\begin{equation}
\mathcal{I}_1=\phi(\mathcal{M}_1)=\frac{exp(\mathcal{M}_1)}{\sum_{n=1}^{N} exp(\mathcal{M}_n)}
\label{eq:conf_map}
\end{equation}

Using the relative importance along with the corresponding embeddings, we create an importance-fused embedding, which is a synthesis of information calculated as follows:

\begin{equation}
\mathcal{F} = \sum_{n=1}^{N} \mathcal{I}_n \odot \mathcal{E}_n \label{eq:conf_estimator}
\end{equation}

\subsection{Temporal Fusion Module}\label{subsec:temporal_module}
To fuse current $\mathcal{E}_t$ and historical $\mathcal{E}_{t-1}, \dots \mathcal{E}_{t-H}$ embeddings with $H$ as the number of historical frames, we use an adapted deformable cross-attention layer \cite{Zhu2021DeformableAttention} from \cite{Yang2023SCOPE}.
Since we only want to focus on the most important features from current and historical frames, we use importance fusion and create an initial fused query embedding.
The initial query is calculated as follows:
\begin{equation}
\mathcal{F}_Q = \mathcal{I}_{t} \odot \mathcal{E}_{t} + \sum_{h=1}^{H} \mathcal{I}_{t-h} \odot \mathcal{E}_{t-h}
\label{eq:deform_query}
\end{equation}
The deformable attention uses reference points $q$ and predicts a 2D spatial offset $\Delta q$ to sample from the keypoints at $q + \Delta q$.
Throughout all experiments, we use a fixed number of 4 simultaneously calculated keypoints ($M = 4)$.
The calculation of the deformable cross-attention at position $q$ can be described as follows:
\begin{equation}
\text{DCM}(q) = \sum_{a=1}^{A} W_a \left[ \sum_{t=-H}^{0} \sum_{m=1}^{M} \phi(W_{b}\mathcal{F}_{Q}(q))\mathcal{E}_{t}(q + \Delta q_m) \right],
\end{equation}
with $a$ as the attention head index, $W_{a/b}$ as the learnable weights and $\phi(\cdot)$ as the softmax function.

\begin{figure*}[ht]
  \centering
  \includegraphics[width=\textwidth]{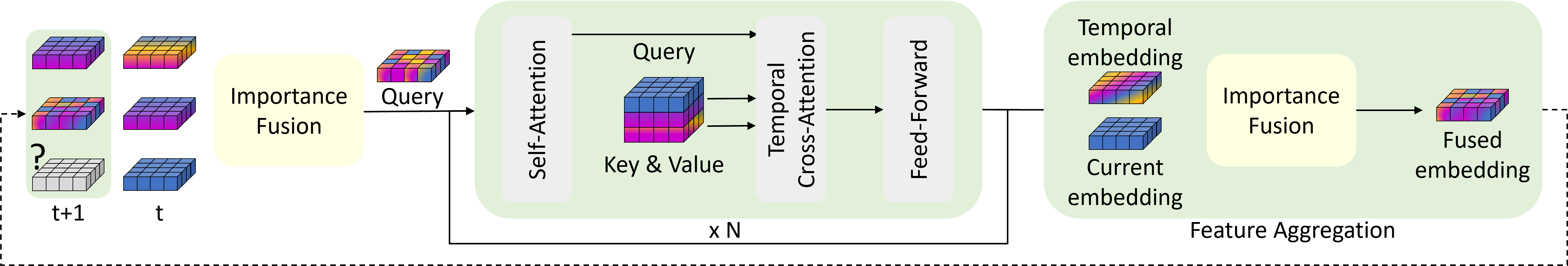}
  \caption{Architecture of TempCoBEV. The question mark and gray embedding refer to the unknown future BEV embedding. TempCoBEV uses historical embeddings, the importance fusion module, and the temporal fusion module to build the embedding for a historical information-integrated BEV prediction.}
  \label{fig:temporal_fusion}
\end{figure*}

We recurrently use $N$ attention blocks for TempCoBEV, each containing self-attention, deformable cross-attention, and a feed-forward layer, following \cite{Li2022BEVFormer, Yang2023SCOPE}.
The overall architecture of the temporal fusion module is shown in Fig. \ref{fig:temporal_fusion}.
In the first iteration, we skip the temporal module and use the output of the original model.
As the temporal fusion module progresses through its $N$ stacks, the output from each iteration serves as the query for the following iteration.
Throughout each stack iteration, the temporal fusion aims to align the historical information with current features, continuously enhancing them with additional historical context.

\subsection{Feature Aggregation}\label{subsec:feature_aggregation}
Considering that the current embeddings represent the information mostly unaffected by temporal distortions \cite{Yang2023SCOPE}, they hold the highest informational significance.
Therefore, we use the importance estimator to assess the certainty of the temporal fusion output and fuse it with the current BEV embeddings, as depicted in Fig. \ref{fig:temporal_fusion}.
Let $\mathcal{I}_T$ be the relative importance of the temporal fusion embedding $\mathcal{E}_T$ and $\mathcal{I}_t$ be the relative importance of the current embedding $\mathcal{E}_t$, then the output based on formula \ref{eq:conf_importance} and \ref{eq:conf_map} is calculated as follows:
\begin{equation}
    \mathcal{F}_{out} = \mathcal{I}_t \odot \mathcal{E}_t + \mathcal{I}_T \odot \mathcal{E}_T
\end{equation}
This method ensures current features are preserved and retrieved in the temporal fusion module, enhancing their contribution to the overall output's informational value.

\section{Experiments}
\subsection{Dataset and Evaluation Metrics}
To assess the effectiveness of the suggested temporal module designed for collaborative perception, we use the OPV2V \cite{Xu2022OPV2V} dataset.
OPV2V is the only publicly available dataset encompassing camera data dedicated to collaborative perception in traffic scenarios and is generated in CARLA \cite{Dosovitskiy2017Carla}.
It offers 73 scenarios, each with an average duration of 25 seconds and featuring a variable count of CAVs, ranging from 2 to 7 per timestamp.
Each CAV has four cameras (front, rear, left, right) to achieve a 360° horizontal field of view with slight overlaps between neighboring cameras.
In the evaluation, we employ ground truth maps to compare non-temporal models and models assisted by temporal information.
Following \cite{xu2022cobevt}, we use the Intersection over Union (IoU) between the predicted and ground truth maps as the evaluation metric, as it assesses the accuracy of object boundaries in BEV map segmentation, a key aspect for tasks related to autonomous driving.
We randomly select an ego vehicle during training for each scenario and use a fixed ego vehicle during testing.
To assess our module, we compare the IoU of the original models with the TempCoBEV extension under two conditions: first, when all historical frames are correct and free from failures, and second, when the subsequent frames have no communication with any other CAV.

\subsection{Implementation Details}
We adopt the configurations from \cite{xu2022cobevt} for all models, which include a communication range of 70m between a CAV and the ego vehicle.
Additionally, we utilize the Adam optimizer and implement a cosine annealing learning rate scheduler.
The model structure remains consistent across all variants, using ResNet34 as the imaging backbone and a decoder with three upsampling layers.
Each upsampling layer includes a bilinear interpolation, 3x3 convolution, batch normalization, and ReLU activation function.
For the temporal fusion module, we use three attention blocks.
We follow the training principle of \cite{Li2022BEVFormer} and use four frames as a sequence, iterating through the frames $t-3, t-2, t-1$ in inference mode, and perform the last step for predicting frame $t$ with gradient calculation.
We employ an approximation of the IoU as the loss function and conduct training for all temporal modules over 40 epochs, utilizing a batch size of 8.
The IoU loss $\mathcal{L}_{\text{IoU}}$ is calculated as follows:
\begin{equation}
    \text{Intersection}(i, j) = \mathcal{D}_{i, j} \cdot \mathcal{G}_{i, j}
\end{equation}
\begin{equation}
    \text{Union}(i, j) = \mathcal{D}_{i, j} + \mathcal{G}_{i, j} - \text{Intersection}(i, j) 
\end{equation}
\begin{equation}
    \mathcal{L}_{\text{IoU}} = 1 - \frac{1 + \sum_{i}^{I}\sum_{j}^{J} \text{Intersection}(i, j)}{1 + \sum_{i}^{I}\sum_{j}^{J} \text{Union}(i, j)}
\end{equation}

The Intersection represents the overlapping area between the predicted \(\mathcal{D}\) and the ground truth segmentation \(\mathcal{G}\) at a specific pixel \((i, j)\) with image size \(I \times J\).
Before applying \(\mathcal{D}\) to the loss, we calculate model output probabilities using the sigmoid function \(\sigma(\mathcal{D}_{i, j})\).
The Union accounts for the total area covered by the predicted and ground truth segmentations, considering overlapping and non-overlapping regions.

The best model selection is determined based on the IoU calculated on the evaluation dataset.
Since we provide a universal temporal module that can seamlessly integrate with all cooperative perception models, our focus remains on retaining the knowledge gained from models' (pre-)training on non-temporal information.
To achieve this, we pre-generate BEV embeddings for each model and entirely remove the original model backbones to only train the temporal module.
We utilize the cooperative BEV embeddings of a model as input while maintaining the original decoder (frozen), leaving the parameters of the pre-trained model unchanged.
The substantial reduction in computational complexity significantly decreases training time, plummeting from approximately 50 minutes to a mere 3 minutes per epoch when executed on an NVIDIA A6000 GPU.

\subsection{Quantitative Evaluation}
\textbf{Current frame prediction.} To evaluate the current frame prediction of the TempCoBEV extension, we use different base models for cooperative driving, specifically, CoBEVT \cite{xu2022cobevt}, F-Cooper \cite{Chen2019FCooper}, V2VNet \cite{Wang2020V2VNet}, and DiscoNet \cite{Li2021DiscoNet}.
We incorporate TempCoBEV into all models without altering their core components or weights.
As illustrated in Figure \ref{fig:model_comparison}, when TempCoBEV is applied, all models either exhibit improved performance or perform comparably to the base models in the current frame prediction.
Notably, the enhancement in IoU is higher for models that initially demonstrate better performance.
We increase the IoU in the current frame predictions of the aforementioned models by up to 2\%.
\\
\textbf{Future frame predictions \& Communication failures.}
In communication failures, data loss from other CAVs necessitates a complete dependence on the ego vehicle's data. 
To assess this scenario, we utilize a sequence comprising at least two historical frames, each featuring one or more additional CAVs.
The objective is to predict the upcoming four future frames solely using the ego vehicle's data and historical cues as inputs.
As depicted in Fig. \ref{fig:model_comparison}, our TempCoBEV mitigates the immediate decline in IoU and outperforms the scenario where only the ego vehicle's input, without historical cues, is utilized for all four future frames.
For CoBEVT, we enhance the IoU for the subsequent future frames by 19\%, 13.1\%, 7.9\%, and 3\%.
For F-Cooper, the improvements are 17.9\%, 11.6\%, 5.7\%, and 1.2\%
For V2VNet by 10\%, 5.8\%, 2.5\%, and 0.2\%.
For DiscoNet by 7.7\%, 3.7\%, 2\%, and 1.1\%.

\begin{figure}
    \centering
    \begin{tikzpicture}
        \begin{axis}[
            xlabel={Time Steps},
            ylabel={IoU},
            xtick={1,2,3,4,5},
            xticklabels={$t$, $t+1$, $t+2$, $t+3$, $t+4$},
            ytick={10,20,...,70},
            legend style={at={(0.8,1.0)},anchor=north},
            grid=both
        ]
        
        \addplot[color=blue,mark=none, solid] coordinates {
            (1,62.42) (2,56.77) (3,50.85) (4,45.67) (5,40.8)
        };
        \addlegendentry{CoBEVT}
        
        \addplot[color=red,mark=none, solid] coordinates {
            (1,46) (2,40.2) (3,33.9) (4,28) (5,23.5)
        };
        \addlegendentry{F-Cooper}

        \addplot[color=green,mark=none, solid] coordinates {
            (1,48.2) (2,44.8) (3,40.6) (4,37.3) (5,35)
        };
        \addlegendentry{V2VNet}

        \addplot[color=orange,mark=none, solid] coordinates {
            (1,48.3) (2,42.4) (3,38.4) (4,36.7) (5,35.8)
        };
        \addlegendentry{DiscoNet}

        \addplot[color=blue,mark=none, dashed] coordinates {
            (1,60.4) (2,37.8) (3,37.8) (4,37.8) (5,37.8)
        };

        \addplot[color=red,mark=none, dashed] coordinates {
            (1,46) (2,22.3) (3,22.3) (4,22.3) (5,22.3)
        };
        \addplot[color=green,mark=none, dashed] coordinates {
            (1,48.2) (2,34.8) (3,34.8) (4,34.8) (5,34.8)
        };
        
        \addplot[color=orange,mark=none, dashed] coordinates {
            (1,48.2) (2,34.7) (3,34.7) (4,34.7) (5,34.7)
        };
        
        \end{axis}
    \end{tikzpicture}
    \caption{Comparison of IoU for different models over time. Solid lines indicate the extension with TempCoBEV. Dashed lines indicate the default model. The dashed green and orange lines overlap heavily.}
    \label{fig:model_comparison}
\end{figure}

\subsection{Qualitative Evaluation}
\begin{figure*}
    \centering
    \begin{tabular}{c}
        CoBEVT (ego-perspective)\\
         \begin{minipage}{0.18\textwidth}
            \centering
            \includegraphics[width=2.6cm, height=2.6cm]{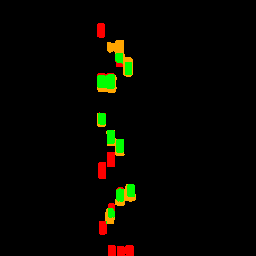}
        \end{minipage}
        \hfill
        \begin{minipage}{0.18\textwidth}
            \centering
           \includegraphics[width=2.6cm, height=2.6cm]{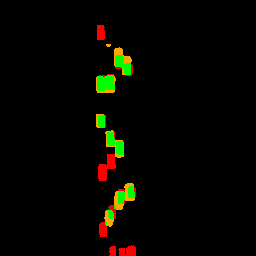}
        \end{minipage}
        \hfill
        \begin{minipage}{0.18\textwidth}
            \centering
            \includegraphics[width=2.6cm, height=2.6cm]{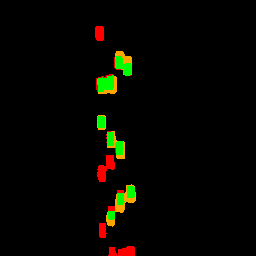}
        \end{minipage}
        \hfill
        \begin{minipage}{0.18\textwidth}
            \centering
            \includegraphics[width=2.6cm, height=2.6cm]{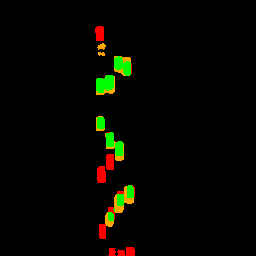}
        \end{minipage}
        \hfill
        \begin{minipage}{0.18\textwidth}
            \centering
            \includegraphics[width=2.6cm, height=2.6cm]{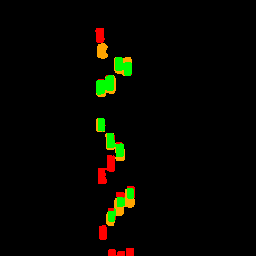}
        \end{minipage}
        \hfill
    \end{tabular}

    \vspace{4pt}

    \begin{tabular}{c}
         CoBEVT (ego-perspective) + TempCoBEV\\
         \begin{minipage}{0.18\textwidth}
            \centering
            \includegraphics[width=2.6cm, height=2.6cm]{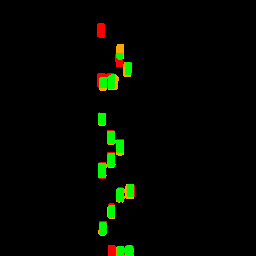}\\
            $t$
        \end{minipage}
        \hfill
        \begin{minipage}{0.18\textwidth}
            \centering
            \includegraphics[width=2.6cm, height=2.6cm]{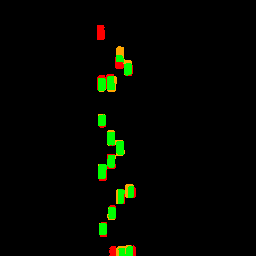}\\
            $t+1$
        \end{minipage}
        \hfill
        \begin{minipage}{0.18\textwidth}
            \centering
            \includegraphics[width=2.6cm, height=2.6cm]{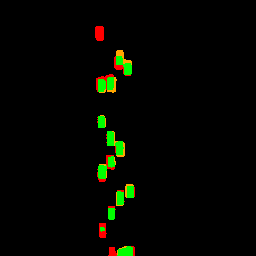}\\
            $t+2$
        \end{minipage}
        \hfill
        \begin{minipage}{0.18\textwidth}
            \centering
            \includegraphics[width=2.6cm, height=2.6cm]{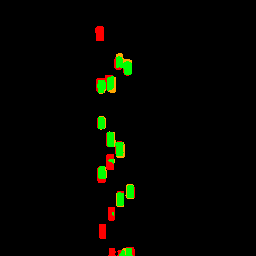}\\
            $t+3$
        \end{minipage}
        \hfill
        \begin{minipage}{0.18\textwidth}
            \centering
            \includegraphics[width=2.6cm, height=2.6cm]{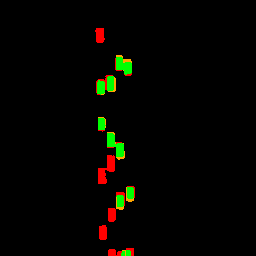}\\
            $t+4$
        \end{minipage}
        \hfill
    \end{tabular}

    \begin{minipage}{\textwidth}
        \centering
        \includegraphics[width=0.95\textwidth]{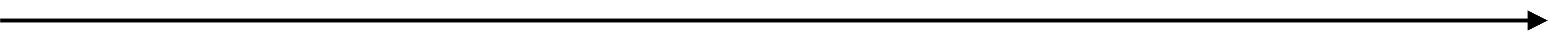}
    \end{minipage}

    \caption{Exemplary output visualization with communication failures (from $t+1$). The first row shows CoBEVT outputs. The second row shows CoBEVT paired with historical cues from TempCoBEV. Green markups are true positives. Red markups refer to false negatives. Orange markups are false positives.}
    \label{fig:qualitiatve_results}
\end{figure*}

Figure \ref{fig:qualitiatve_results} illustrates the qualitative outcomes of TempCoBEV integrated with the optimal base model, CoBEVT.
As outlined in Section \ref{subsec:feature_aggregation}, the significance of importance-fusion in the feature aggregation component is emphasized, leading to a higher weighting of current frame features over historical ones over time.
However, TempCoBEV adeptly leverages historical cues, seamlessly incorporating them into subsequent frame predictions to enhance current and future frames, particularly in communication failure scenarios.

\subsection{Ablation Studies}
\textbf{Component analysis.} The significance of the Importance Estimator is shown in Table \ref{tbl:component_analysis}, illustrating its importance for both the attention mechanism query and feature aggregation.
We use CoBEVT \cite{xu2022cobevt} as the model for comparison.
The collaborative integration of these components significantly contributes to substantially enhancing overall performance.

\textbf{Historical Frames for Fusion.} As TempCoBEV employs a recurrent approach to optimize BEV embeddings, the inclusion of additional historical embeddings ($\geq 2)$ results in a performance decline, shown in Table \ref{tbl:ablation_findings}.
We believe that employing more than one historical embedding enhances the risk of error accumulation and predicts wider regions of potential vehicle positions, leading to a decrease in the IoU.

\textbf{Data Augmentation.} Including additional parameters in a model can potentially increase the risk of overfitting.
In the case of the temporal module, vulnerability to overfitting and the potential to recover historical cues depend on data augmentation.
Table \ref{tbl:ablation_findings} compares model performances with and without data augmentation, emphasizing the necessity for greater data diversity to improve overall performance.

\textbf{Temporal Fusion.} When comparing with the straightforward utilization of the mean of historical frames, the effectiveness of TempCoBEV is demonstrated in Table \ref{tbl:ablation_findings}.
It recovers historical cues and indicates the capacity to learn and anticipate the dynamics of moving objects.

\textbf{Loss Function.} Even though the CoBEVT architecture employs a weighted cross-entropy loss function for training (and performs suboptimally with the IoU loss), TempCoBEV demonstrates improved performance using the IoU loss.
The performance disparities between the two loss functions are depicted in Table \ref{tbl:ablation_findings}, highlighting the effectiveness of the IoU loss, particularly for future frame predictions.

\begin{table}
    \caption{Component analysis (IoU). TCB = TempCoBEV, FA = Feature Aggregation, QI = Query Importance}\label{tbl:component_analysis}
    \begin{center}
        \begin{tabular}{ccccccccc}
            \hline
            \textbf{TCB} & \textbf{FA} & \textbf{QI} & $t$ & $t+1$ & $t+2$ & $t+3$ & $t+4$ \\ \hline
            $\times$ & N.A. & N.A. & 60.4 & 37.8 & 37.8 & 37.8 & 37.8 \\
            \checkmark & $\times$ & $\times$ & 59.7 & 53.9 & 47.7 & 42.3 & 37.8 \\
            \checkmark & \checkmark & $\times$ & 57.3 & 49.9 & 43.4 & 37.4 & 32.6\\
            \checkmark & $\times$ & \checkmark & 60.8 & 54.0 & 47.6 & 42.1 & 38.0\\
            \checkmark & \checkmark & \checkmark & \textbf{62.4} & \textbf{56.8} & \textbf{50.8} & \textbf{45.7} & \textbf{40.8} \\ \hline
        \end{tabular}
    \end{center}
\end{table}

\begin{table}
    \centering
    \caption{Ablation study findings to training designs and strategies. ``Comm'', ``Aug'', ``Flip'', and ``Rot'' means Communication, Augmentation, Flipping, and Rotation, respectively.}
    \label{tbl:ablation_findings}
    \begin{tabular}{c|ccccc}
        \hline
        Design Strategies & $t$ & $t+1$ & $t+2$ & $t+3$ & $t+4$ \\ \hline \hline
        Full model & \textbf{62.4} & \textbf{56.8} & \textbf{50.8} & \textbf{45.7} & \textbf{40.8} \\ \hline
        \multicolumn{6}{c}{Effect of Historical Frames for Fusion} \\ \hline
        1 Frame (Default) & $62.4$ & $56.8$ & $50.8$ & $45.7$ & $40.8$ \\
        2 Frames & $61.9$ & $55.3$ & $45.3$ & $39.3$ & $36.9$ \\
        3 Frames & $59.1$ & $52.6$ & $45.9$ & $40.1$ & $36.4$ \\
        4 Frames & $60.5$ & $55.4$ & $47.6$ & $39.9$ & $34.7$ \\
        \hline
        \multicolumn{6}{c}{Effect of Data Augmentation} \\ \hline
        No Aug. & $62.7$ & $38.1$ & $36.1$ & $36.1$ & $36.1$ \\
        Comm. Aug. (Default) & $62.4$ & $56.8$ & $50.8$ & $45.7$ & $40.8$ \\
        Comm. Aug. + Flip & $62.3$ & $54.9$ & $47.3$ & $40.9$ & $38.2$ \\
        Comm. Aug. + Flip + Rot & $62.3$ & $54.6$ & $47.4$ & $41.7$ & $38.9$ \\ \hline
        \multicolumn{6}{c}{Effect of Temporal Fusion} \\ \hline
        Mean & $60.4$ & $54.1$ & $45.7$ & $39.3$ & $34.9$ \\
        TempCoBEV (Default) & $62.4$ & $56.8$ & $50.8$ & $45.7$ & $40.8$ \\ \hline
        \multicolumn{6}{c}{Effect of Loss Function} \\ \hline
        Weighted Cross-Entropy & $57.6$ & $43.7$ & $37.0$ & $28.9$ & $24.2$ \\
        IoU (Default) & $62.4$ & $56.8$ & $50.8$ & $45.7$ & $40.8$ \\ \hline
        \hline
    \end{tabular}
\end{table}

\section{Conclusion and Limitations}
This paper proposes an independent temporal module explicitly designed for cooperative BEV map segmentation.
Our module uses importance-guided attention, focusing on current and historical embeddings during fusion.
Through extensive experiments, we show that TempCoBEV surpasses single-frame cooperative perception models in communication failure scenarios and exhibits superior performance in current and future BEV map predictions.
Given the absence of publicly available real multi-vehicle camera-based datasets, we utilize the OPV2V dataset.
Consequently, we train and evaluate all models on synthetic data without explicitly modeling realistic V2V challenges.
Future research extends the applicability to more realistic and complex perception scenarios.

\section*{Acknowledgment}
This work was partially funded by the Bavarian state government as part of the High Tech Agenda and BayWISS.


\end{document}